%% file: main.tex
\documentclass[conference]{IEEEtran}
\IEEEoverridecommandlockouts

\input{header}

\begin{document}

\title{Gaussian Mixture Graphical Lasso with \\ Application to Edge Detection in Brain Networks}

\author{\IEEEauthorblockN{1\textsuperscript{st} Hang Yin}
\IEEEauthorblockA{\textit{Data Science} \\
\textit{Worcester Polytechnic Institute}\\
Worcester, USA \\
hyin@wpi.edu}
\and
\IEEEauthorblockN{2\textsuperscript{nd} Xinyue Liu}
\IEEEauthorblockA{\textit{Alexa AI} \\
\textit{Amazon}\\
Boston, USA \\
cinvro@gmail.com
}
\and
\IEEEauthorblockN{3\textsuperscript{rd} Xiangnan Kong}
\IEEEauthorblockA{\textit{Computer Science} \\
\textit{Worcester Polytechnic Institute}\\
Worcester, USA \\
xkong@wpi.edu}
}

\maketitle

\begin{abstract}
\small\baselineskip=9pt
Sparse inverse covariance estimation (\textit{i.e.}, edge detection)
is an important research problem in recent years, where the goal is to discover the direct connections between a set of nodes in a networked system based upon the observed node activities. 
Existing works mainly focus on unimodal distributions, where it is usually assumed that the observed activities are generated from a \emph{single} Gaussian distribution (\textit{i.e.}, one graph).
However, this assumption is too strong for many real-world applications.
In many real-world applications (\textit{e.g.}, brain networks), the node activities usually exhibit much more complex patterns that are difficult to be captured by one single Gaussian distribution.
In this work, we are inspired by Latent Dirichlet Allocation (LDA) \cite{blei2003latent} and consider modeling the edge detection problem as estimating a mixture of \emph{multiple} Gaussian distributions, where each corresponds to a separate sub-network.
To address this problem, we propose a novel model called Gaussian Mixture Graphical Lasso (MGL). 
It learns the proportions of signals generated by each mixture component and their parameters iteratively via an EM framework.
To obtain more interpretable networks, MGL imposes a special regularization, called Mutual Exclusivity Regularization (MER), to minimize the overlap between different sub-networks.
MER also addresses the common issues in read-world data sets, \textit{i.e.}, noisy observations and small sample size. 
Through the extensive experiments on synthetic and real brain data sets, the results demonstrate that MGL can effectively discover multiple connectivity structures  from the observed node activities.

\end{abstract}


\section{Introduction}

Edge detection of brain network \cite{fabijanska2008edge} aims at identifying the edges between nodes (\textit{i.e.}, functionally coherent brain regions) of a brain mapping \cite{belliveau1995fmri} from a temporal sequence of observed activities (\textit{e.g.}, fMRI scans).
Since a well-constructed connectivity network servers as the prerequisite for many graph mining algorithms on brain disorder diagnosis and brain functionality analysis \cite{ahmadlou2012graph}, it is significant to design a more effective and accurate edge detection method.
\input{butterfly.tex}
Existing edge detection methods usually rely on the assumption that all nodes' activities obey a multivariate Gaussian distribution, and the connections between nodes could be depicted by their inverse covariance matrix (\textit{a.k.a.} precision matrix).
A widely used variation of this line of works is known as Graphical Lasso (GLasso) \cite{friedman2008sparse}, which additionally imposes sparseness on the precision matrix.
However, in many neurology studies such as \cite{diez2018neurogenetic}, human brains usually exhibit dramatically different activity modes when they perform different tasks.
Based on these studies, we believe that the cognitive structure of the human mind can be paralleled into several sub-graphs based on different cognitive control processes and behavior. 
Cognitive control means a set of dynamic processes that engage and disengage different nodes of brain to modulate attention and switch between tasks.
Applying GLasso without considering different latent cognitive modes is equivalent to deriving an ``average'' network representation.
Since the behavior of different brain modes varies significantly, the derived ``average'' network may lose crucial information.
Under such context, as illustrated in Figure~\ref{fig:butterfly}, it is natural to investigate whether and how one could extend the edge detection methods applied in brain network to capture the connectivity structures of multiple underlying cognitive brain modes.

\input{family.tex}

To incorporate the concept of multiple connectivity structure into edge detection, we follow the idea of latent Dirichlet allocation (LDA) \cite{blei2003latent} to adopt Gaussian mixture model on this problem.
LDA views a document as a mixture of various topics, and it assumes that the generation of a document follows some topic-word distributions which can be found by sampling. 
Similarly, we could view brain scans as mixtures of latent modes, where each mode is characterized by a Gaussian distribution with different covariance $\Sigma_{\text{M}}$.
Each covariance matrix $\Sigma_{\text{M}}$ corresponds to a specific connectivity among brain nodes. 
In the generation of each brain node activity, our model chooses a mode $\text{M}$ based on the mode distribution $\pi$ (as LDA chooses a topic), and then it generates a brain node activity $A_{i} \sim \text{Multinomial}(\mathbf{0},\Sigma_{\text{M}})$ (as LDA generates a word based on the topic chosen).
Figure~\ref{fig:family} illustrates the relations and differences between our proposals and LDA, we also compare with traditional edge detection methods Graphical Lasso~\cite{friedman2008sparse}, where all brain node activities are assumed being produced by a single unified zero-mean $\Sigma$-covariance multivariate Gaussian distribution.  

In this paper, our goal is to reveal these structure of underlying sub-network from the observed activities simultaneously.
To solve above issues, our main challenges are as follows:
\begin{itemize}
    \item \textbf{Mixture of multiple connectivity networks}: 
    In real-world cases, the proportions and assignments of each mode are not observable. Without the prior knowledge of them, general GLasso only discovers a simple graph for the whole data sets.
    While our problem setting requires estimating the proportions and assignments of multiple latent cognitive modes as well as the parameters of the network for each mode, with the same input as GLasso, which is much more challenging.

    \item \textbf{Direct connectivity among the nodes}: 
    The finite Gaussian Mixture Model (GMM) \cite{pearson1894contributions}  seems a straightforward solution to our problem, which incorporates a heterogeneous structure into the graphical model. 
    It fits multivariate normal distributions and treats proportions and assignments as prior and posterior probabilities (estimators) in the Bayesian setting respectively. However, it estimates the covariance of each distribution rather than the inverse covariance, which indicates that the discovered connections could be indirect and make the network unnecessarily complicated.
    So GMM is inappropriate to distinguish the directed relationships between each pair of nodes. 
    
    \item \textbf{Noisy Observations and Small Sample}:
    It is already a challenging task to discover a single network given the noisy observations and the small size of the data sample. GLasso employs simple $\ell_1$-norm regularization on to alleviate the sensitiveness to noises, but it is not sufficient for our case.
    Based on the cognitive studies on the human brain \cite{lin2017dynamic}, each brain sub-network is not only sparse but also has limited overlapping with other sub-network, simply adopting $\ell_1$-norm regularization as in GLasso may make the derived sub-network highly intertwined and hard to interpret.
    So we want to design a new regularization into the model, which can enforce it to discover a set of different sub-graphs, no matter small sample size or noisy data.
\end{itemize}

To tackle the above challenges, we propose a new model, namely MGL, to discover such mixture connectivity structures of the brain network.
Similar to GMM, MGL learns the proportions and assignments of each latent cognitive mode iteratively via an EM framework, with the emphases on inferring the inverse covariance matrix of each latent distribution.
A novel regularization approach called Mutual Exclusivity Regularization (MER) is also proposed to differ each inverse covariance matrix, implying that sub-network of different brain regions are activated under different cognitive modes.

\section{Preliminary}
\subsection{Notation}
Throughout this paper, $\mathbb{R}$ denotes the set of all real numbers, $\mathbb{R}^n$ stands for the n-dimensional euclidean space.
The set of all $m \times n$ matrices with real entries is denoted as $\mathbb{R}^{m \times n}$.
All matrices are written in boldface.
We write $\mb{X} \succ 0$ to denote that matrix $\mb{X}$ is positive definite. 
We write $\text{tr}(\cdot)$ to refer the trace of a matrix, which is defined to be the sum of the elements on the main diagonal of the matrix.
We use $\vert \mb{X} \vert$ to denote the determinant of a real square matrix $\mb{X}$. 
We define a special matrix of $\mb{X}$ as follows:
\begin{equation}
\begin{aligned}
    \bar{\mb{X}} = \begin{bmatrix} 0 & |X_{12}|  &\cdots &|X_{1N}|\\|X_{21}|&0&\cdots&|X_{2N}| \\ |X_{N1}| &|X_{N2}|&\cdots &0\end{bmatrix} 
\end{aligned}
\end{equation}
$\bar{\mb{X}}$ is the non-negative copy of $\mb{X}$ removed all diagonal elements.

\subsection{Graphical Lasso}
Graphical Lasso (GLasso) or Gaussian Graphical Model (GGM) is usually formulated as the following optimization problem,
\begin{equation}
\begin{aligned}
    \label{eq:glasso}
	\min_{\mb{\Theta} \succ 0} -\text{log} \vert \mb{\Theta} \vert + \text{tr} (\mb{S} \mb{\Theta}) + \lambda ||\mb{\Theta}||_1\\
\end{aligned}
\end{equation}
where $\mb{S} = \frac1n \mb{X}^{\top} \mb{X}$ is the empirical covariance matrix. 
$\mb\Theta = \mb\Sigma^{-1}$ is defined as the inverse covariance matrix, which can filter the directed links between all relationships.
$||\mb{\Theta}||_1$ is the $\ell_1$-norm regularization that encourages sparse solutions, and $\lambda$ is a positive parameter denotes the strength of regularization.

\section{MGL Method}
\subsection{Gaussian Mixture Graphical Lasso}
Given the number of base distributions $K$ and the number of node $N$, we assume the observed sample of each node is a mixture of the $K$ distributions. 
Thus, the joint probability of all observations $\mb{X} = (\bsym{x}_1^\top, \cdots, \bsym{x}_N^\top) \in \mathbb{R}^{N \times D}$ is given by
\begin{align*}
    \label{eq:mggl}
	  p(\mb{X} \vert \mb{\Theta}_k,  \bsym{\mu}_k, \phi_k) &= \prod_{i=1}^{N}\sum_{k=1}^{K}  \phi_k \mc{N}(\bsym{x}_i \vert \bsym{\mu}_k, \mb{\Sigma}_k) 
\end{align*}
We could assume $\bsym\mu_k = \bsym{0}$ without losing generality, so the negative log likelihood (NLL) in terms of $\{\mb{\Theta}_k\}$ is given by,
\begin{equation}
\label{eq:mgglsimple}
\begin{aligned}
	 \text{NLL}(\bsym\theta) 
	 = - \sum_{i=1}^{N}\text{log}\Big(\sum_{k=1}^{K} \phi_k\mc{N}(\bsym{x}_i \vert \bsym{0}, \mb{\Theta}_k^{-1})\Big)
\end{aligned}
\end{equation}
where $\bsym\theta = \{\phi_1, \cdots, \phi_k, \mb{\Theta}_1, \cdots \mb{\Theta}_k\}$ is the model parameters.

\subsection{The Mutual Exclusivity Regularization}
Similar to the Adaptive Lasso in cite{zou2006adaptive}, we also need to impose regularization on our mixture model to obtain interpretable results, which means non overlapping edges exist among all estimators of precision matrices. 
However, be different with adaptive lasso or fused lasso, the intuitions are two folds: (1) we want each $\mb{\Theta}_k$ to be sparse; (2) we want each $\mb{\Theta}_k$ to be fairly different from other $\mb{\Theta}_{k'}$.
Towards this end, we propose to the mutual exclusivity regularization as follows,
\begin{align}
   \ell_{\lambda_1, \lambda_2}(\{\mb{\Theta}_k\}) = \lambda_1 \sum_{k=1}^{K} \Vert \mb{\Theta}_k \Vert_1 + \lambda_2 \sum_{i\ne j} \text{tr}(\bar{\mb{\Theta}}_i \bar{\mb{\Theta}}_j)
\end{align}
where $\bar{\mb{\Theta}}$ is the non-negative copy of $\mb{\Theta}$ removed all diagonal elements. 
The first term is identical to graphical lasso, which imposes sparsity controlled by $\lambda_1 > 0$ on each $\mb{\Theta}_k$.
The second term is the summation of the approximate divergence measure between each pair $(\mb{\Theta}_i,\mb{\Theta}_j)$.
It is easy to see when there is no overlapping non-zero entities between each $\mb{\Theta}_k$, this term reaches its minimal value $0$.  
$\lambda_2 > 0$ is employed to tune the strength of the second regularization. So it makes sense that we can use this term to force each estimation of $\mb{\Theta}_k$ in the result to have as few over-lapping elements as possible.

Hence, we formally present the objective of our MGL as follows,
\begin{align}
    \min_{\{\mb{\Theta}_k \succ 0\}} \text{NLL}(\{\mb{\Theta}_k\}) +  \ell_{\lambda_1, \lambda_2}(\{\mb{\Theta}_k\})
\end{align}

\subsection{The Latent States} Since there are $K$ separate latent distributions, so each data sample $\bsym{x}_i$ could come from one of the $K$ distributions, we denote the corresponding state as $z_{i} \in \{1,\cdots, K\}$.  
Thus, the NLL function could be rewritten as follows,
\begin{equation}
\begin{aligned}
\label{latent}
    \text{NLL}(\bsym\theta) &= - \sum_{i=1}^{N}\log\sum_{k=1}^{K}\Big(\frac{\mb{Q}(z_{ik}) p\big(\bsym{x}_i \vert  \mb{\Theta}_k\, \phi_k\big)}{\mb{Q}(z_{ik})}\Big)\\
    &= - \sum_{i=1}^{N}\log\sum_{k=1}^{K}\Big(\frac{p\big(\bsym{x}_i,z_{ik} \vert \mb{\Theta}_k\, \phi_k\big)}{\mb{Q}(z_{ik})}\Big)
\end{aligned}
\end{equation}
Here $\mb{Q}(z_{ik})$ is the latent variable and $\sum_{k=1}^{K} \mb{Q}(z_{ik}) = 1 $. 

According to the expression in the Equation (\ref{latent}), it can not be directly computed because the expression in $log$ is a sum term. So we can use the Expectation Maximization (EM) algorithm to optimize the above NLL \emph{w.r.t.} $\{\mb{\Theta}_k\}$. We summarized the MGL algorithm in Algorithm (\ref{mggl}).

\begin{algorithm}[t]
\caption{Algorithm for \texttt{MGL}}
\label{mggl}
\begin{algorithmic}[1]
\Require 

i: $\mb{X}$: The observations of $D$-variate Gaussian distribution

ii: $k$: the number of Gaussian distributions

iii: $\lambda_1$: the Lagrangian multiplier of sparsity constraint

iv: $\lambda_2$: The Lagrangian multiplier of mutual exclusivity constraint

v: $\text{iter}_{max}$: the maximum number of iteration

Output: $\mb{\hat{\Theta}_k}$, $\mb{\Hat{\phi}_k}$
\State Initialization: initialize $\mb{\phi}_k^{(0)}$, $\mb{\Theta}_k^{(0)}$ and $r_{ik}^{(0)}$
\Repeat
\State E step: Update the latent variable $r_{ik}^{(t)}$ with given $\mb{\phi}_k^{(t-1)}$ and $\mb{\Theta}_k^{(t-1)}$
\State M step: Update $\mb{\phi}_k^{(t)}$, $\mb{\Theta}_k^{(t)}$ with  $r_{ik}^{(t-1)}$
\Until{$iter={iter}_{max} \text{ or convergence}$}
\end{algorithmic}
\end{algorithm}

\input{f1.tex}

\subsection{Initialization}

As we know from the Algorithm (\ref{mggl}), we need to give starting values of each estimators. 
In the process of comparative experiments, we found that the initialization of the parameters will largely affects the performance of our model.
The following scheme we found empirically works well in our experiments. For each observation $i = 1,\dots,N$, we distribute it randomly a class $k \in \{1,\dots,K\}$. Then we assign a weight $\hat{r}_{ik} = 0.9$ for this observation $i$ and distribution $k$ and $\hat{r}_{ij} = \frac{0.1}{K-1}$ for all other distributions. In the M-step, we update $\mb{\Theta}_k$ from the initial values $\hat{\mb{\Theta}}_k^{(0)}$ computed by GLasso based on the whole samples. and $\phi_k$ from the initial values $\hat{\phi}_k = \frac{1}{K}$.

\section{Empirical Study}

In this part, we demonstrate the performance of our proposed model through extensive comparative experiments.
We evaluate our proposed model in synthetic datasets at first. To comprehensively evaluate proposed model, we conduct experiment to answer the following research questions:
\begin{itemize}
\item{\textbf{RQ 1}:} How does MGL perform compared with state-of-the-art models in the consideration of the effect of sample size?
\item{\textbf{RQ 2}:} Does our model still show robustness under noise? If the MER regularization term has positive influence on the performance under noise?
\item{\textbf{RQ 3}:} How do hyper-parameters in comparative experiments impact each model performance?
\item{\textbf{RQ 4}:} Is there a problem with mixture brain network structure in real ADHD-200 datasets? 
\end{itemize}

\subsection{Compared Baselines}
To demonstrate the effectiveness of our proposed method, we test against several variations of the state-of-art method Graphical Lasso:
\begin{itemize}
    \item \textbf{GLasso + Spectral Clustering}: GLasso algorithm that assumes all data samples are drawn from the same Gaussian distribution, then using Spectral Clustering divide the whole network into several sub-graph.
    \item \textbf{\emph{k}-means + GLasso}: This is a pipeline method that first employs \emph{k}-means to assign each $\bsym{x}_i$ to different groups, then  using GLasso for each group to obtain the final $\mb{\Theta}_k$.
    \item \textbf{JGL \cite{gao2016estimation}}: This is the Joint Graphical Model with fused lasso, which is proposed in \cite{gao2016estimation}. It is equivalent to our proposed model without MER term. So it can work as the comparative method for assessing the performance of MER. 
\end{itemize}

\subsection{Synthetic Simulations}
Due to the lack of ground truth in many real-world data, we first compare our proposed method against other competitors on several carefully designed synthetic data sets.

\subsubsection{Data Set}
In this sub-section, we design some synthetic data sets purposefully. Firstly, we generate $k$ diagonal matrices ($k$ is the number of distribution, which is given in advance), then divide it into several equal-scale blocks. It makes sense for two reasons: we need to control each sub-graphs $\mb{\Theta}_k$ with non-overlapping edges on off-diagonal areas; by making edges of each sub-graph more concentrated, it is helpful for making results conductive to visualization. Secondly, we choose different off-diagonal blocks on each $\mb{\Theta}_k$, giving connectivities for these chosen blocks with a high density. Following the above steps, we generate each $\mb{\Theta}_k$ without overlapping edges on off-diagonal areas. Based on $\mb{\Theta}_k$, we compute each $\mb{\Sigma}_k$, then select $N_k$ samples ($\sum_{k=1}^{K} N_k = N$) randomly from each Gaussian distribution. In the next subsection, in order to evaluate the stability of our model, we also add noise into the samples. To exclusive the system randomness, we sample 10 times for all experiments, calculate the average of each experiments. So we can evaluate the precision and stability of our model at the same time.

\subsubsection{Experimental Settings}
We simulate four scenarios by controlling one parameter and holding on the others. In these situations, we select sample size $N$ and the standard error of noise $\mb{\sigma}$ as the controlled parameters. 
\begin{itemize}
    \item \textbf{Scenario 1}: We fix $p=8$ (the number of variables), $k=2$ (the number of Gaussian distributions), and $\mb{\sigma}=0$ (the standard error of noise), and then control sample size $N$ from 100 to 520. 
    \item \textbf{Scenario 2}: We fix $p=8$, $k=2$, and $N=500$, and then control noise $\mb{\sigma}$ from 0.1 to 0.8. 
    \item \textbf{Scenario 3}: We fix $p=20$, $k=2$, and $\mb{\sigma}=0$, and then control sample size $N$ from 200 to 1000. 
    \item \textbf{Scenario 3}: We fix $p=20$, $k=2$, and $N=1000$, and then control noise $\mb{\sigma}$ from 0.1 to 0.8. 
\end{itemize}

\subsubsection{Evaluation}
To evaluate the quality of each sub-graph, we define the F1-score of edge detection as $F1 = \frac{2 N_d^2}{N_a N_d + N_g N_d}$, where $N_d$ is the number of true edges detected by the model, $N_g$ is the number of true edges and $N_a$ is the total number of edges detected. According to the expression, higher F1-score indicates better quality of edge detection.

Figure \ref{fig:f1-score} shows the comparison between MGL and other baseline models. The results in the figure answer the first three \textbf{RQ} mentioned at the beginning of this section. The first column shows the results when we control sample size $N$ and hold on the others, which corresponds to \textbf{RQ1}. It is obvious that $k$-means and Spectral models are useless when the ground truth data sets are drawn from mixture Gaussian distribution. Meanwhile, when the sample size is not large enough, the precision of JGL is lower than that with MGL. The second column shows the results when we control noise, which corresponds to \textbf{RQ2}. We fix the sample size $N$ on 500, so when $\mb{\sigma}=0$, JGL is as good as MGL. According to the results, The louder the noise, the worse JGL performs, which means sensitive to the noise.So the result demonstrates that MER regularization can improve the performance of our proposed model. Compared to the others, MGL shows robustness in this scenario. To answer \textbf{RQ3}, we can figure out the answer from both column in this figure.
Since our experiments are setting in low-dimensional and high-dimensional space separately, we can see from all comparison results that the issue of hyper-parameters does not affect the performance of MGL. In contrast, the performance of JGL in high-dimensional space isn't as well as that in the low-dimensional space, no matter in the scenario of sample size or noise.
In summary, in the comparative experiment of synthetic datasets with ground-truth, our proposed method MGL shows better accuracy and robustness than that of other comparison methods.


\subsection{Real fMRI Data}

In the subsection, we evaluate our proposed method on fMRI dataset from ADHD-200 project\footnote{http://fcon\_1000.projects.nitrc.org/indi/adhd200}. 
Through this paper, we discover the network discovery from a collection of fMRI scans, in which each sample corresponds to a 4D brain image (a sequence of 3D images) of a subject. Our real world dataset is distributed by nilearn\footnote{http://nilearn.github.io/}. 
Specifically, there are 40 subjects in total. Among them, 20 subjects are labeled as ADHD, and the others are labeled as TDC. The fMRI scan of each subject in the dataset is a series of snapshots of 3D brain images of size $61 \times 76 \times 61$ over $\sim$176 time steps. 

In our experiment, we only choose the subjects which are labeled as ADHD. We focus on the multiple connectivity structures among the same subjects, in order to provide evidence on feature selection between different subjects in further study. Rather than discover the brain network on the level of voxels, we extracts the signal on regions defined via a probabilistic atlas, to construct the data sets. So it is more conventional for visualization of the results. The data sets is a $1899 \times 39$ data sets and we consider that they are drawn from a mixture Gaussian distribution. However, the number $k$ of it is unknown, which need to be given in advance. Through repeated experimental observations, we found that $k=4$ can provide the most reasonable results on the data sets. 

Because real fMRI data lacks ground-truth as a reference to measure the accuracy and robustness of the model. We are more concerned with the interpretability and rationality of the results. Specific to our proposed model, we are more concerned about whether our model can mine different connectivity structures among nodes from the fMRI datasets. 

\begin{figure}[t]
\centering
\begin{subfigure}{\columnwidth}
\centering
  \includegraphics[width=.2\linewidth]{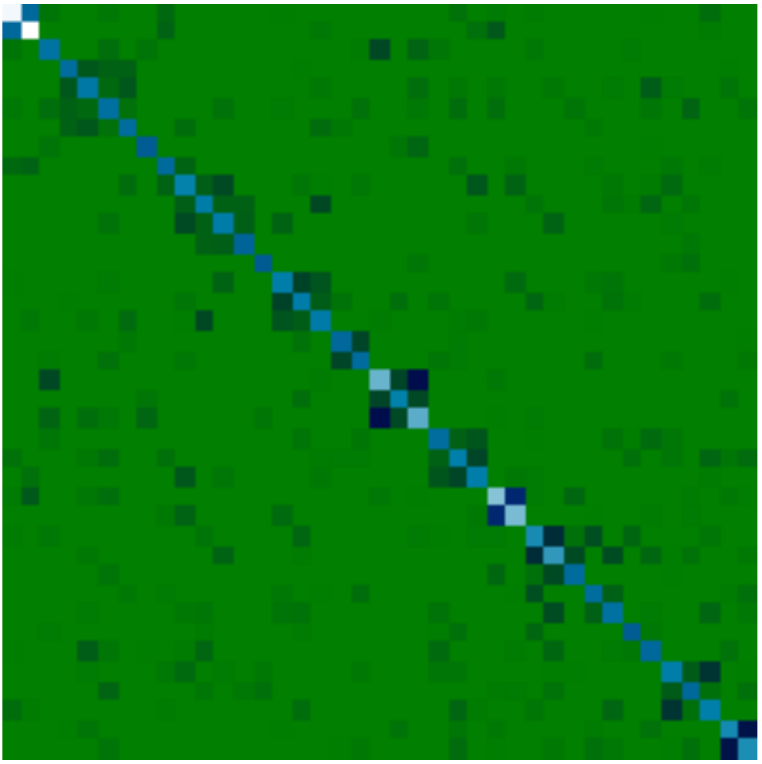}
  \includegraphics[width=.2\linewidth]{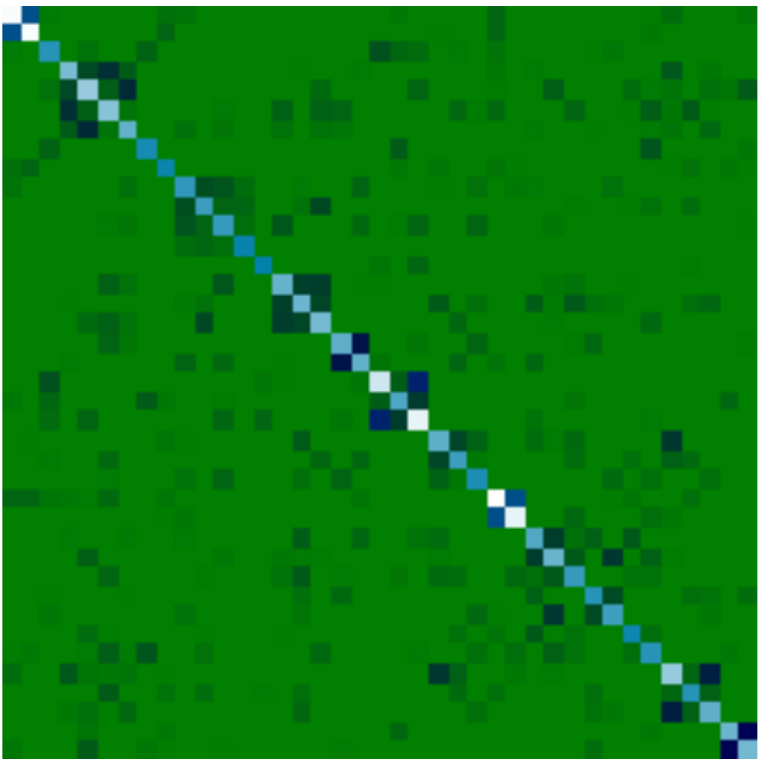}
  \includegraphics[width=.2\linewidth]{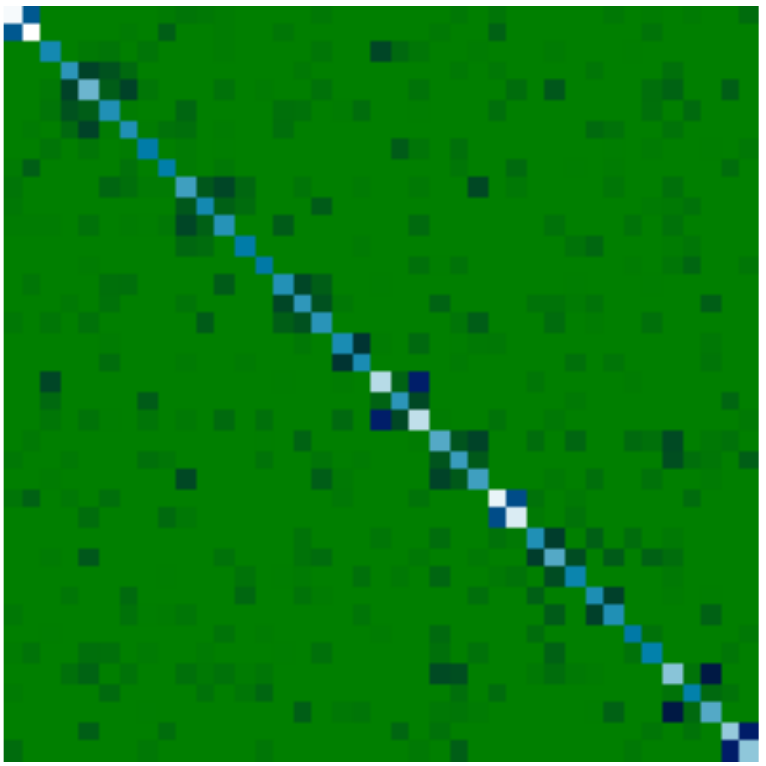}
  \includegraphics[width=.2\linewidth]{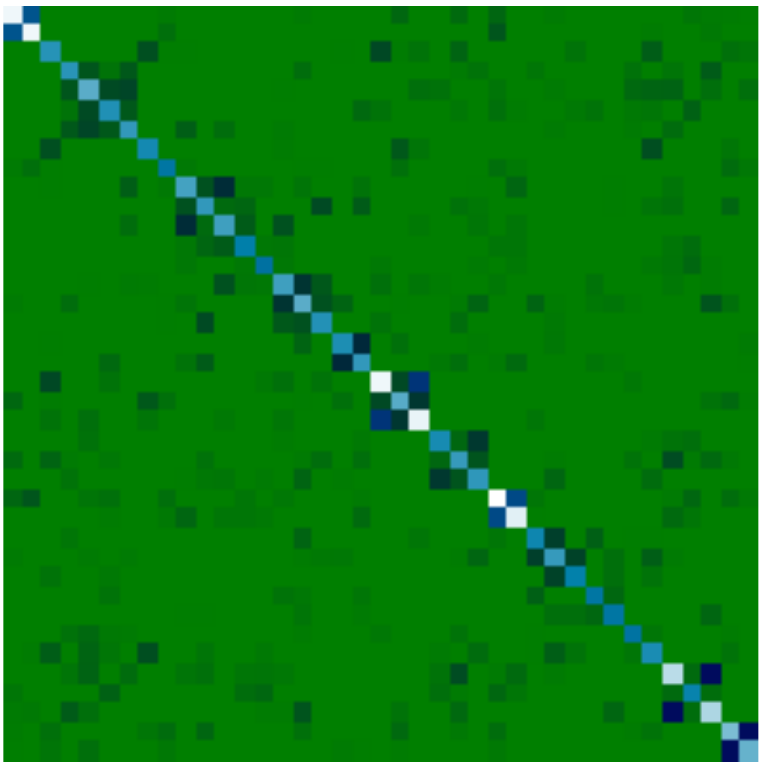}
  \caption{Sub-graphs discovered by $k$-means}
\end{subfigure}
\begin{subfigure}{\columnwidth}
\centering
  \includegraphics[width=.2\linewidth]{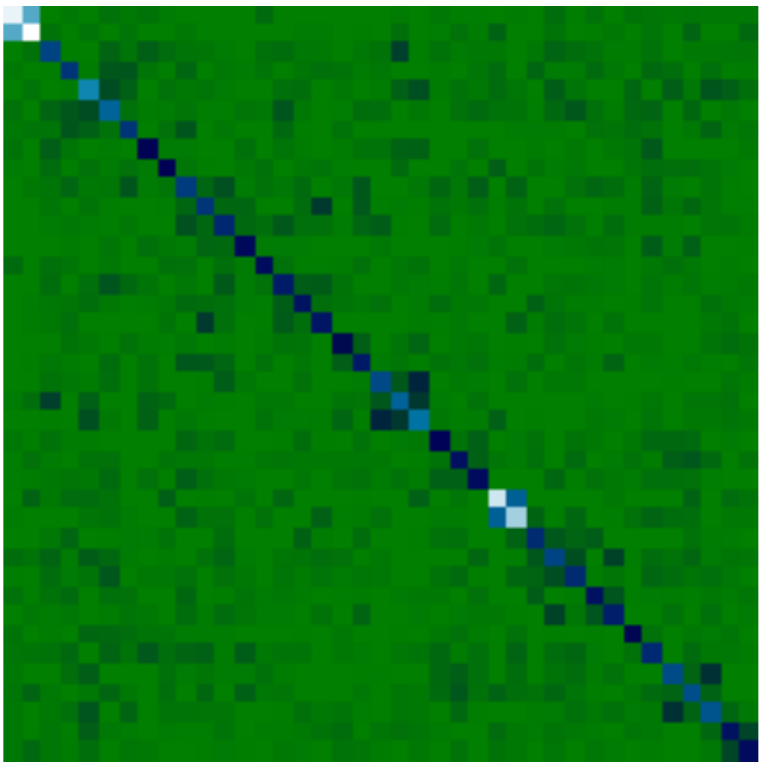}
  \includegraphics[width=.2\linewidth]{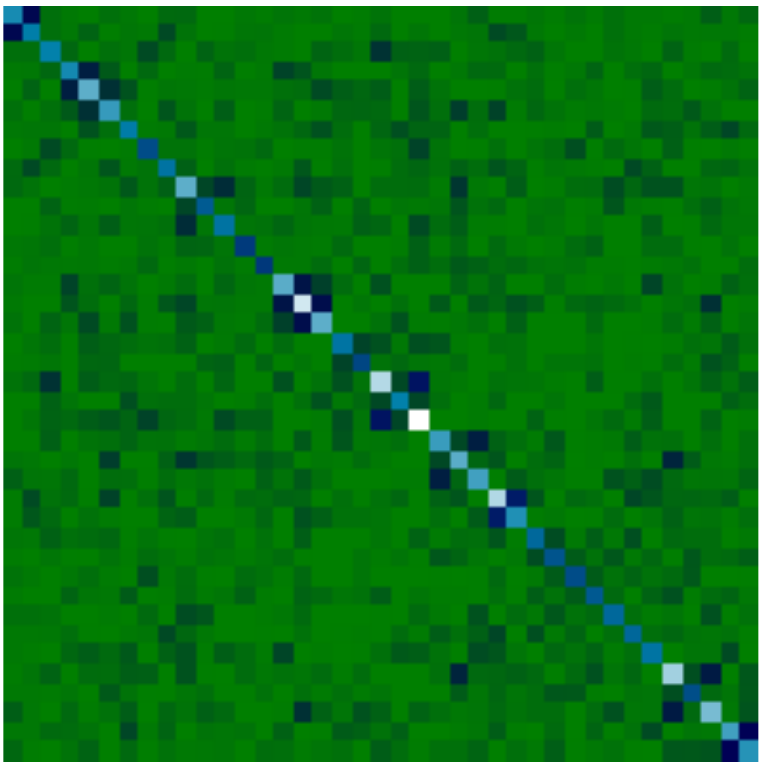}
  \includegraphics[width=.2\linewidth]{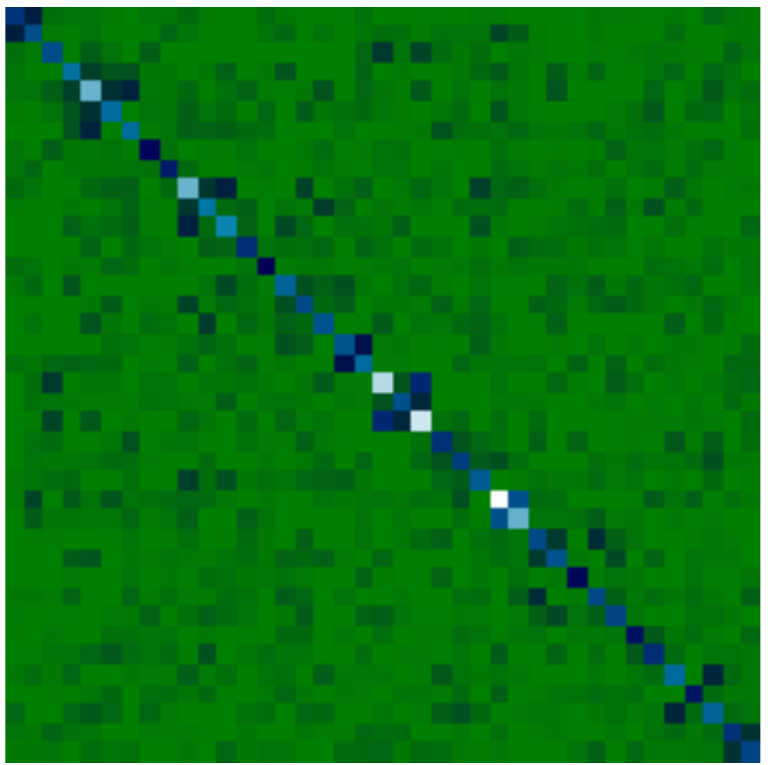}
  \includegraphics[width=.2\linewidth]{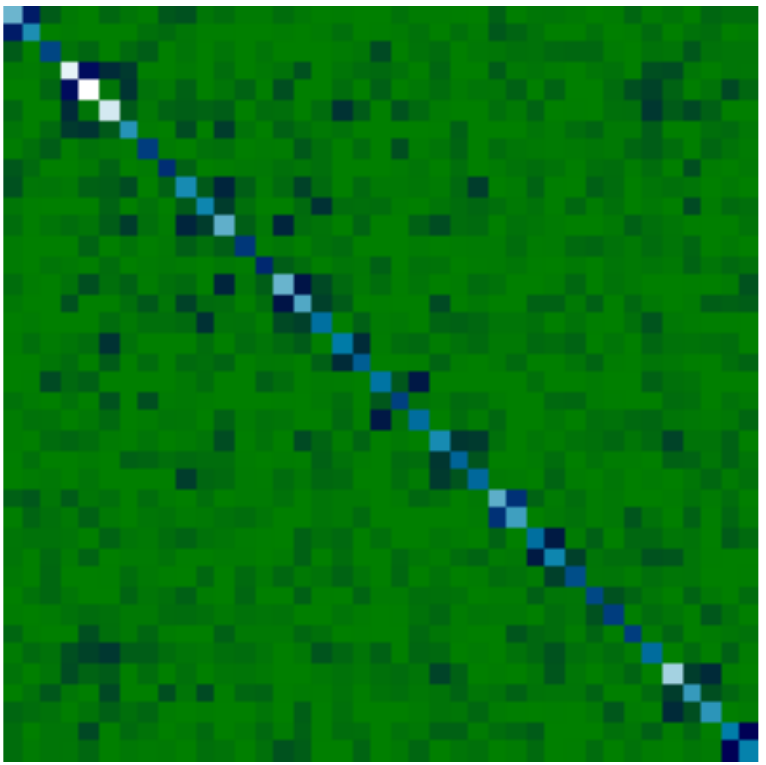}
  \caption{Sub-graphs discovered by JGL}
\end{subfigure}
\begin{subfigure}{\columnwidth}
\centering
  \includegraphics[width=.2\linewidth]{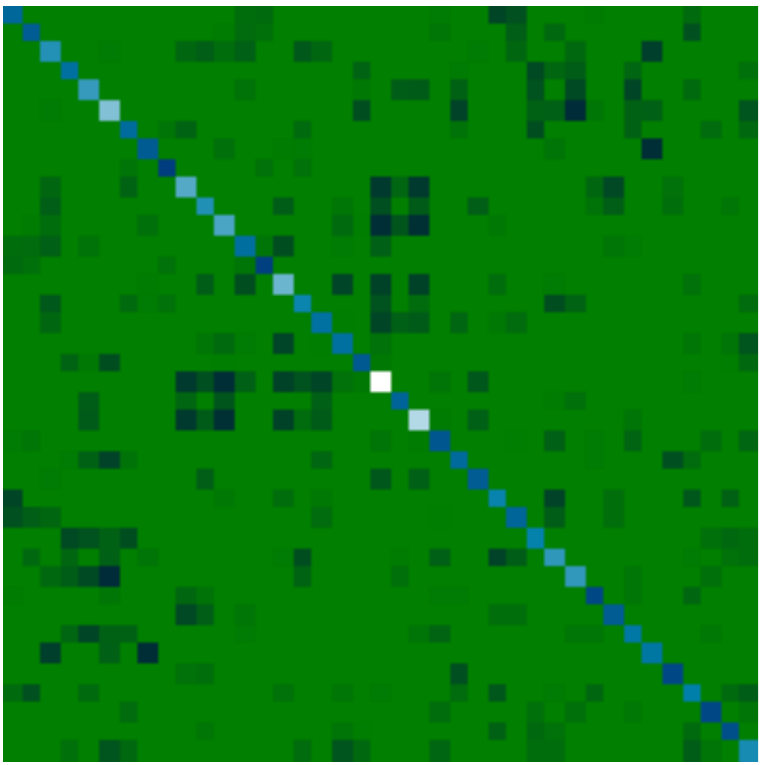}
  \includegraphics[width=.2\linewidth]{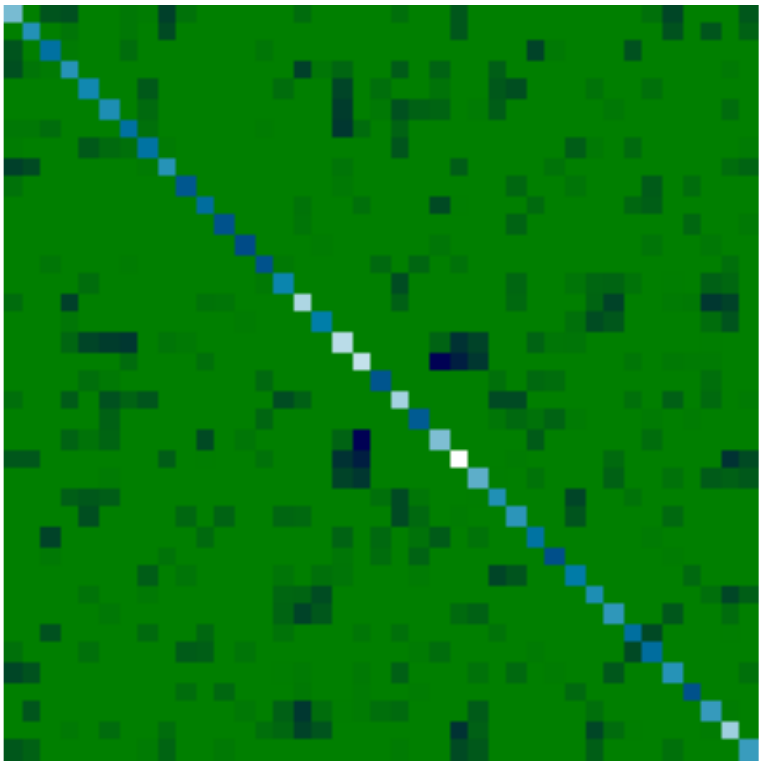}
  \includegraphics[width=.2\linewidth]{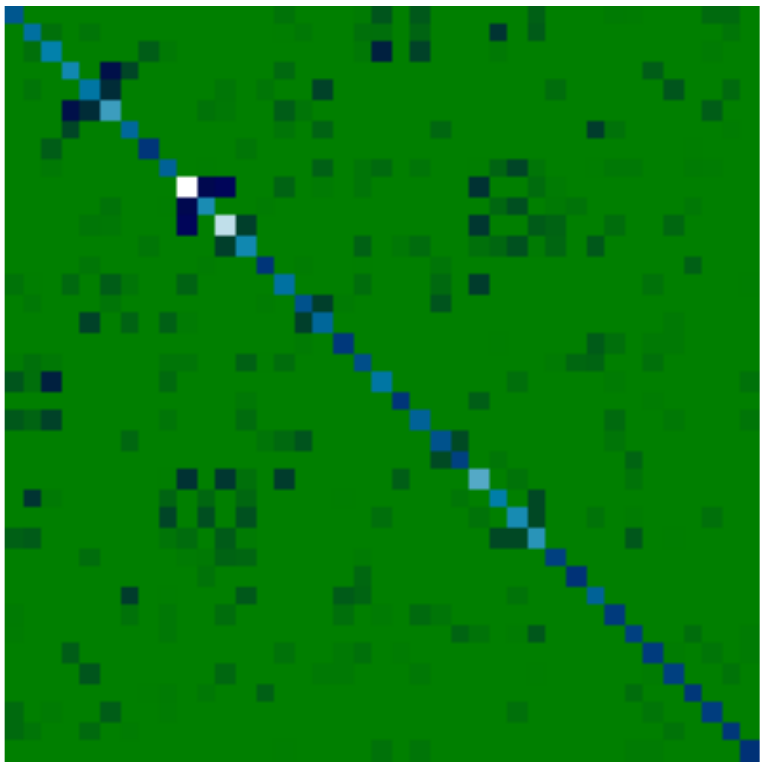}
  \includegraphics[width=.2\linewidth]{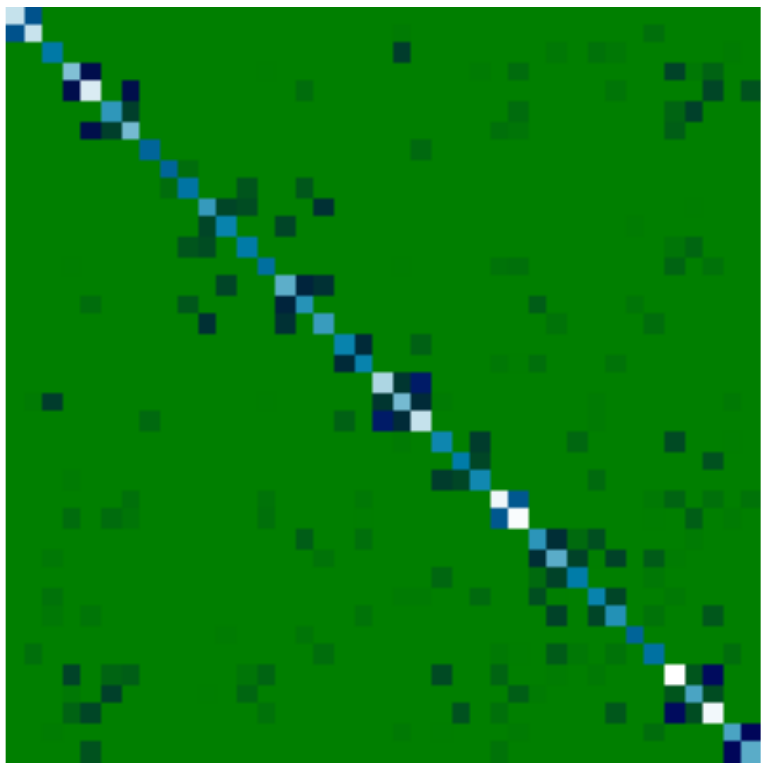}
  \caption{Sub-graphs discovered by MGL}
\end{subfigure}
\caption{Comparison of $k$-means + GLasso, JGL and MGL on ADHD dataset. 
The results show how to estimate a mixture connectivity structure on a group of subjects using different group sparse inverse covariance estimation models from real fMRI data set.
The closer the color of elements in off-diagonal is to blue, the bigger probability the directed edges between corresponding nodes.}
\label{fig:adhd}
\vspace{5pt}
\end{figure}

\begin{figure}[t]
\centering
\begin{subfigure}{0.45\columnwidth}
\centering
  \includegraphics[width=1\linewidth]{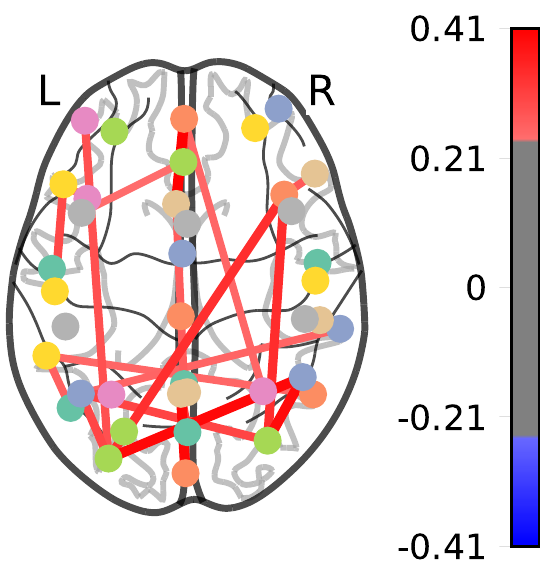}
\end{subfigure}
\begin{subfigure}{0.45\columnwidth}
\centering
  \includegraphics[width=1\linewidth]{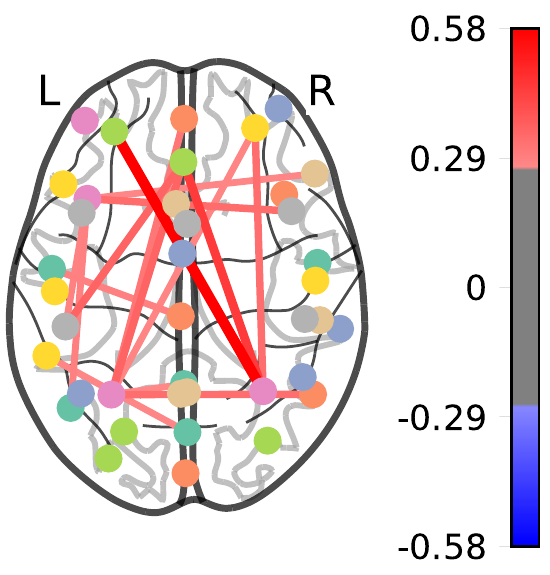}
\end{subfigure}
\begin{subfigure}{0.45\columnwidth}
\centering
  \includegraphics[width=1\linewidth]{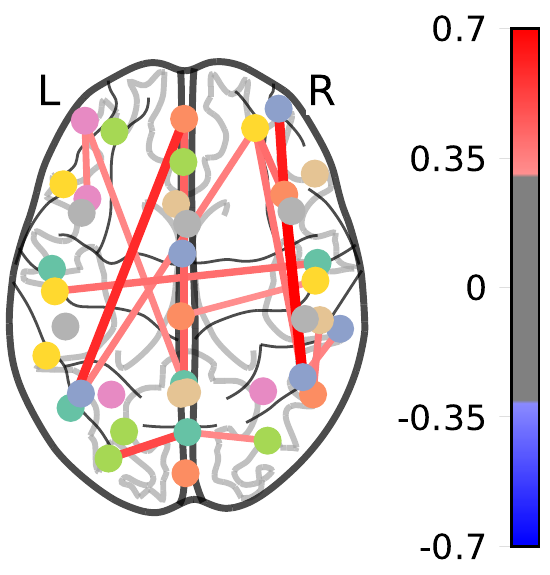}
\end{subfigure}
\begin{subfigure}{0.45\columnwidth}
\centering
  \includegraphics[width=1\linewidth]{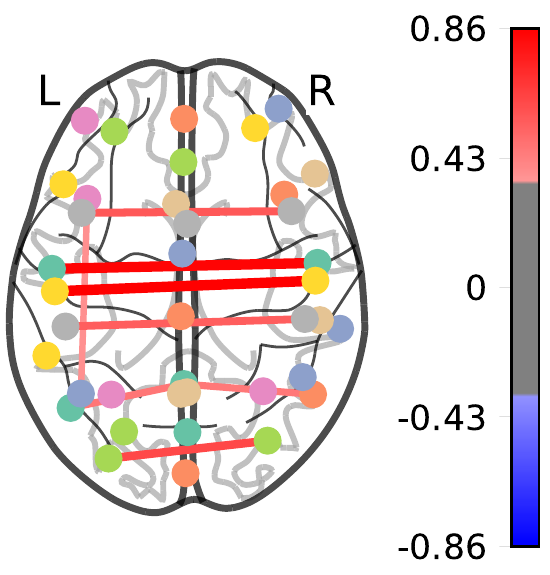}
\end{subfigure}
\caption{
We turn the results of Fig. \ref{fig:adhd} into connectome for visualization. Each precision matrix is displayed on glass brain on extracted coordinates.
These graphs of precision matrices discovered by MGL in ADHD dataset. The closer the color of edge is to red, the stronger the directed relationship between corresponding nodes.}
\label{fig:connectome}
\vspace{5pt}
\end{figure}

According to the Figure \ref{fig:adhd}, we can find that there are almost no differences among four sub-graphs discovered by $k$-means plus GLasso. It indicates that this method is useless for mining sub-graphs in ADHD data sets. JGL shows four different sub-graphs, however, so many overlapped areas among them. 
These results seem not to be sparse matrices, which indicates that the corresponding connectivity structure is not very clear through this method.
Compared to it, sub-graphs discovered by MGL is clearer and the number of overlapped areas is less. 
Therefore, although lacking the ground truth in ADHD data, we can still believe that the inferred results of MGL is consistent with the defined problem in this paper, especially in the consideration of mixture Gaussian distribution with non-overlapping areas among their precision matrices. 
The Figure \ref{fig:connectome} shows the corresponding connectivity structure of the results discovered by MGL. Here we only choose the axial direction of the cuts to show. The closer the color is to red, the stronger the directed relationship between the corresponding nodes. We highlight the stronger edges by adjusting the threshold of colorbar.
According to the visualization of results, we can see that different sub-graphs highlight different relationships among all nodes. Different sub-graphs emphasize the relationships of different nodes, which means that subjects present different network structures on the time-line. This phenomenon is more obvious between the nodes related to DMN (default mode network), which includes the Parietal, Occipital Lobes, the Cingulum Region Posterior and the Frontal Cortex.
Although the hypothesis about non-overlapped areas among each connectivity structure may not exist in real ADHD subjects, we believe that MGL with MER regularization can more prominently show the difference between each connectivity structures discovered, so that we can have a better understanding of the association between cognitive network and human activities.

According to the analysis above, despite the lack of ground-truth, we believe that the existing results are still consistent with the problem defined in this paper. 
So the result shows that there is a mixture connectivity structure among nodes in the fMRI datasets, and our proposed model MGL can effectively mine this mixture connectivity structure.

\section{Related Work}

In the edge detection of brain network, it has two major branches: effective connectivity estimation and functional connectivity estimation. 
For the first branch, scholars pay more attention on obtaining a directed network from fMRI data through structure learning method for Bayesian networks \cite{huang2011brain}. 
In contrast, the second branch focuses on some approaches such as hierarchical clustering, pairwise correlations and independent component analysis, which can be found in \cite{friston2011functional} for more details. 
\cite{friedman2008sparse} proposed sparse gaussian graphic models , which are a very useful for discovering directed links of brain network based on large-scale dataset by using sparse inverse covariance estimation. 
However, in the task of edge detection, these methods focus on unimodal distributions, where it is usually assumed that the observed samples are drawn from a single Gaussian distribution, which is opposed to some recent studies \cite{anderson2018developmental}. 
The Joint Graphical Model with fused lasso, which is proposed in \cite{gao2016estimation}, is in the framework of multivariate Gaussian mixture modeling. However, this method has shown to be sensitive to the noise and small size of the data sample.

\section{Conclusion}

To address the problem of mixture connectivity substructures between nodes in brain network discovery, we propose embedding one of the current methods of estimating multiple Gaussian graphical models in the framework of Gaussian mixture modeling, then design a new regularization term, called mutual exclusivity regularization, to make sub-graphs un-overlapped with each other. 
Through extensive controlled experiments, we demonstrate that our proposed model MGL shows more effectiveness than other baseline models, meanwhile, MGL shows more robustness than JGL, especially in the consideration of small samples or noisy data sets. In addition, this conclusion is also demonstrated in the experiment of real fMRI brain scanning datasets from ADHD subjects. So we have reason to believe that, our method can also be applied in other domains when network connectivity structure is very complex.

\footnotesize{
\balance
\bibliographystyle{plain}
\bibliography{header,reference}}
\end{document}

%% file: header.tex
\usepackage{booktabs,makecell,tabularx}

\usepackage{setspace}

\newcolumntype{L}{>{\raggedright\arraybackslash}X}
\usepackage{siunitx}
\usepackage{adjustbox}
\usepackage{array,booktabs}
\usepackage{graphicx}
\usepackage{epstopdf}
\usepackage{algorithm}
\usepackage{algpseudocode}
\usepackage{multirow}
\usepackage{amsmath}
\usepackage{amsfonts}
\usepackage{amssymb}
\usepackage{color}
\usepackage{enumitem}
\usepackage{subcaption}
\usepackage{balance}

\newcommand{\bsym}{\boldsymbol}
\newcommand{\mb}{\mathbf}
\newcommand{\mc}{\mathcal}

\def\BibTeX{{\rm B\kern-.05em{\sc i\kern-.025em b}\kern-.08em
    T\kern-.1667em\lower.7ex\hbox{E}\kern-.125emX}}

%% file: butterfly.tex
\begin{figure}[t]
\centering
\begin{minipage}{1\columnwidth}
    \includegraphics[width=\textwidth]{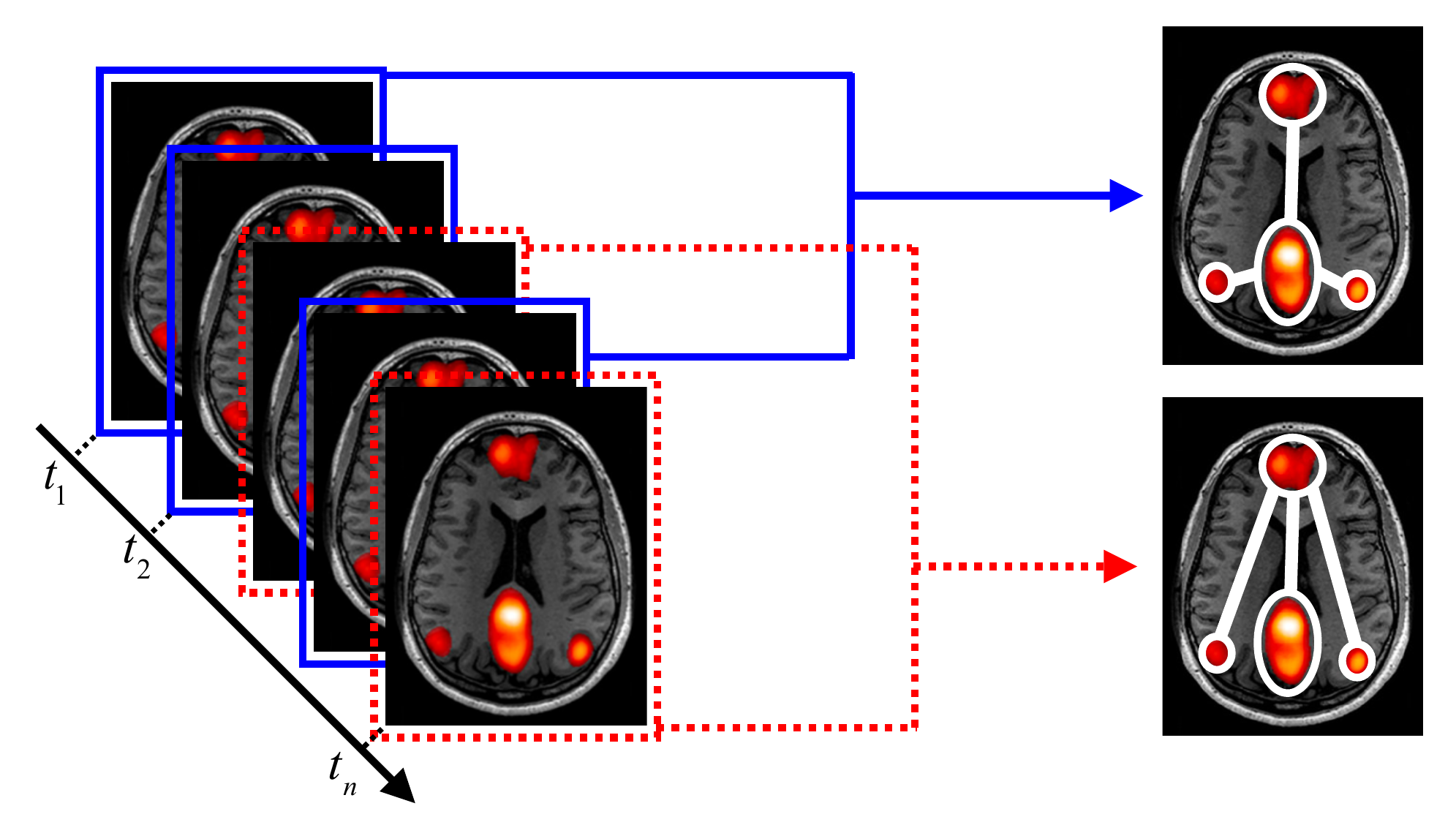}
\end{minipage}
\caption{
    The problem of Gaussian mixture sparse inverse covariance estimation.
    The brain activities over time may originate from the mixture of multiple latent cognitive brain modes (\emph{i.e.} different connectivity structures among nodes).
    Without knowing the mode proportions and assignments in the observed brain images, our goal is to discover these underlying sub-networks for different modes. 
}
\label{fig:butterfly}
\vspace{5pt}
\end{figure}

%% file: family.tex
\begin{figure}[t]
\centering
\begin{subfigure}{1\columnwidth}
    \centering
    \includegraphics[width=\textwidth]{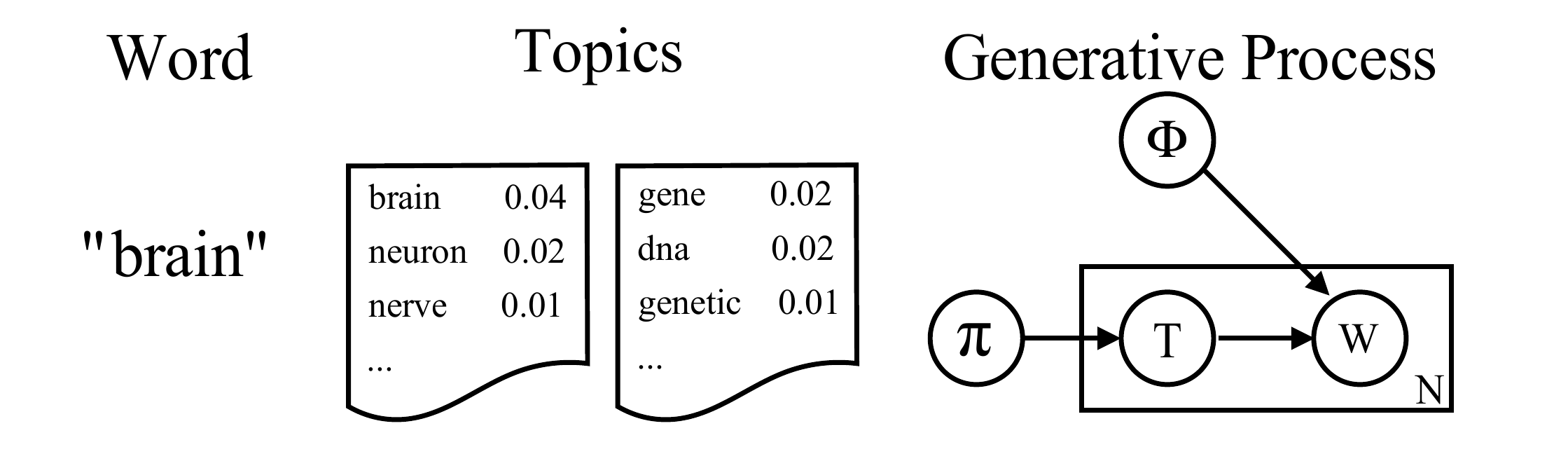}
    \caption{Latent Dirichlet Allocation (LDA)\cite{blei2003latent}}
\end{subfigure}
\begin{subfigure}{1\columnwidth}
    \centering
    \includegraphics[width=\textwidth]{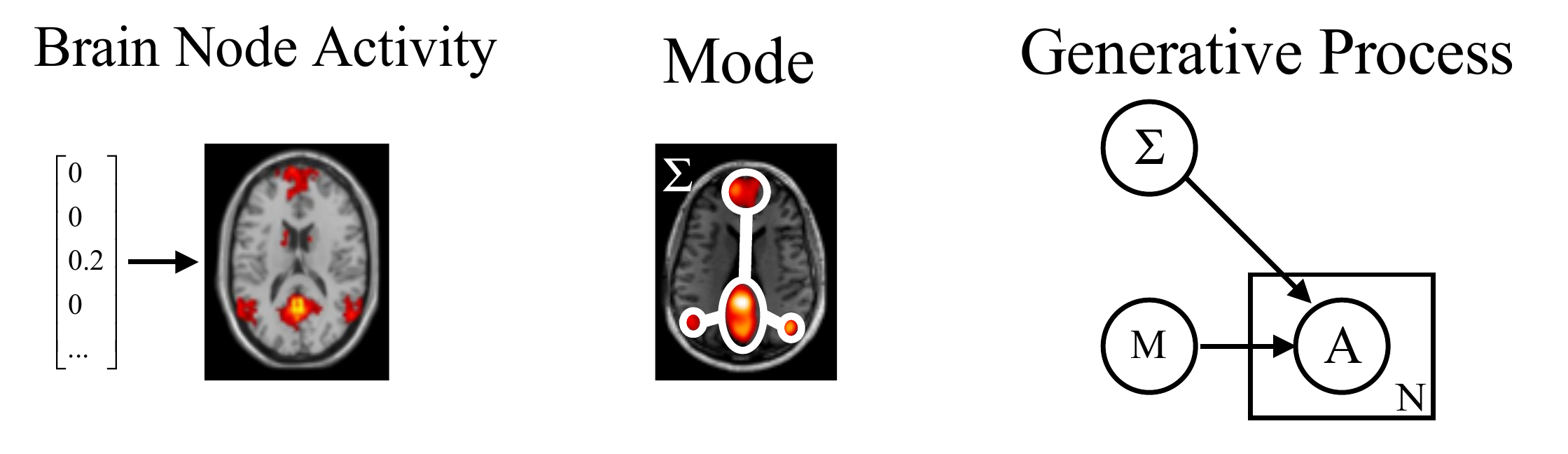}
    \caption{Graphical Lasso\cite{friedman2008sparse}}
\end{subfigure}
\begin{subfigure}{1\columnwidth}
    \centering
    \includegraphics[width=\textwidth]{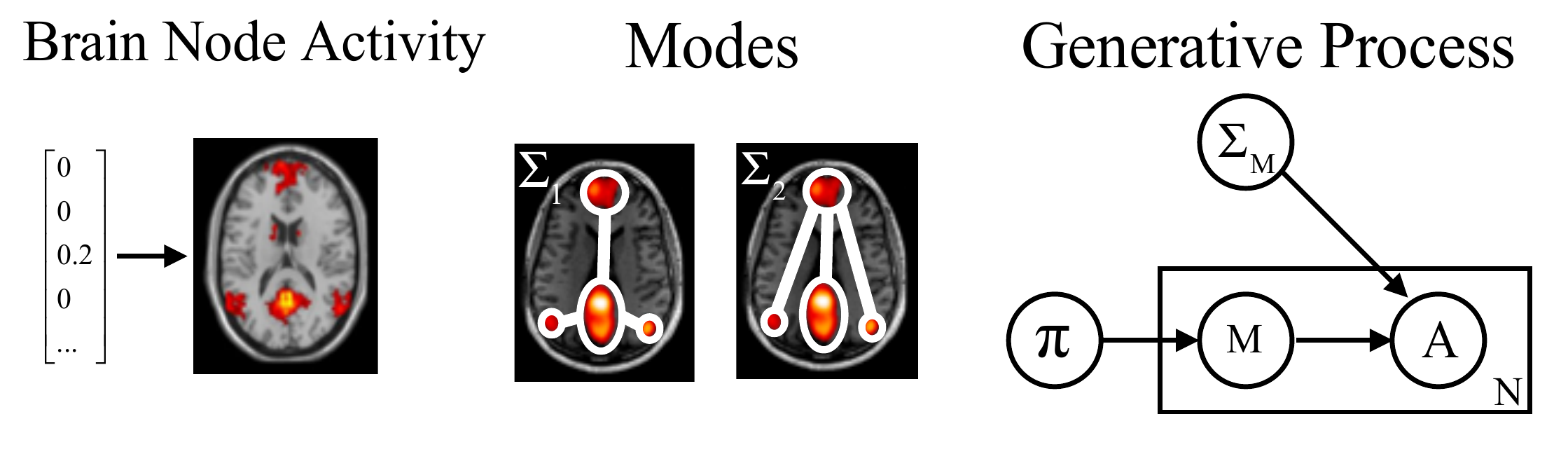}
    \caption{Mixture Graphical Lasso (this paper)}
\end{subfigure}
\caption{Comparison of Latent Dirichlet Allocation(LDA), Graphical Lasso (GLasso) and our model in this paper. In each sub-graph, the boxes are "plates" representing replicates, which are repeated entities. 
The outer plate represents document in LDA or observation subject in brain network study, while the inner plate represents the generative process of word ($W$) in a given document or brain node activity ($A$) in a given subject, each of which word or scan is associated with a choice of topic ($T$) or mode ($M$) and the parameter of the corresponding word or node activity distribution ($\phi$ or $\Sigma$). $\pi$ is the topic or mode distribution. $N$ denotes the number of words or scans.}
\label{fig:family}
\vspace{5pt}
\end{figure}

%% file: f1.tex
\begin{figure}[t]
\centering
\begin{subfigure}{.45\columnwidth}
\centering
  \includegraphics[width=1\linewidth]{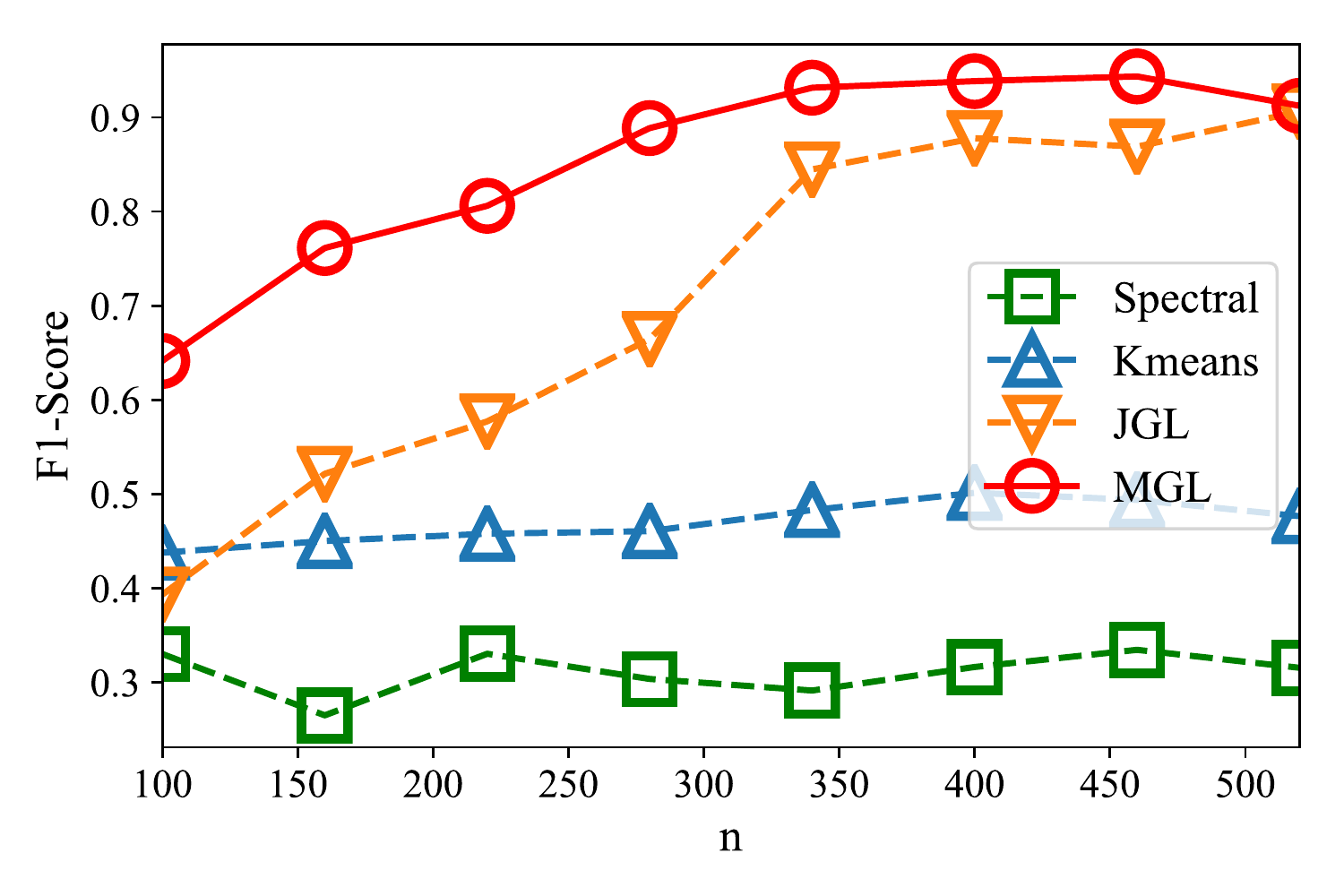}
  \caption{Low Dimension - Sample Size (Scenario 1)}
\end{subfigure}
\begin{subfigure}{.45\columnwidth}
\centering
  \includegraphics[width=1\linewidth]{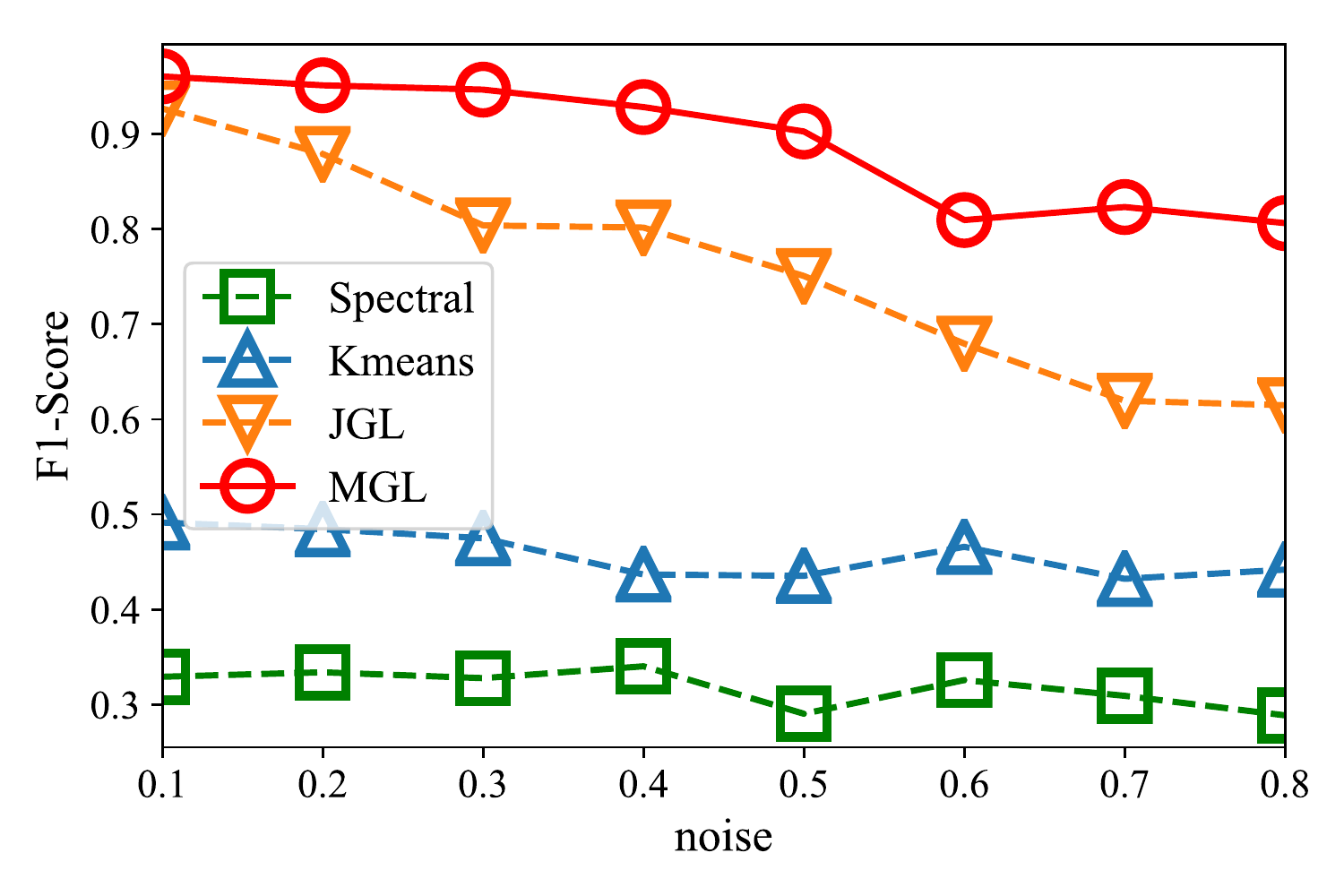}
  \caption{Low Dimension - Noise (Scenario 2)}
\end{subfigure}
\begin{subfigure}{.45\columnwidth}
\centering
  \includegraphics[width=1\linewidth]{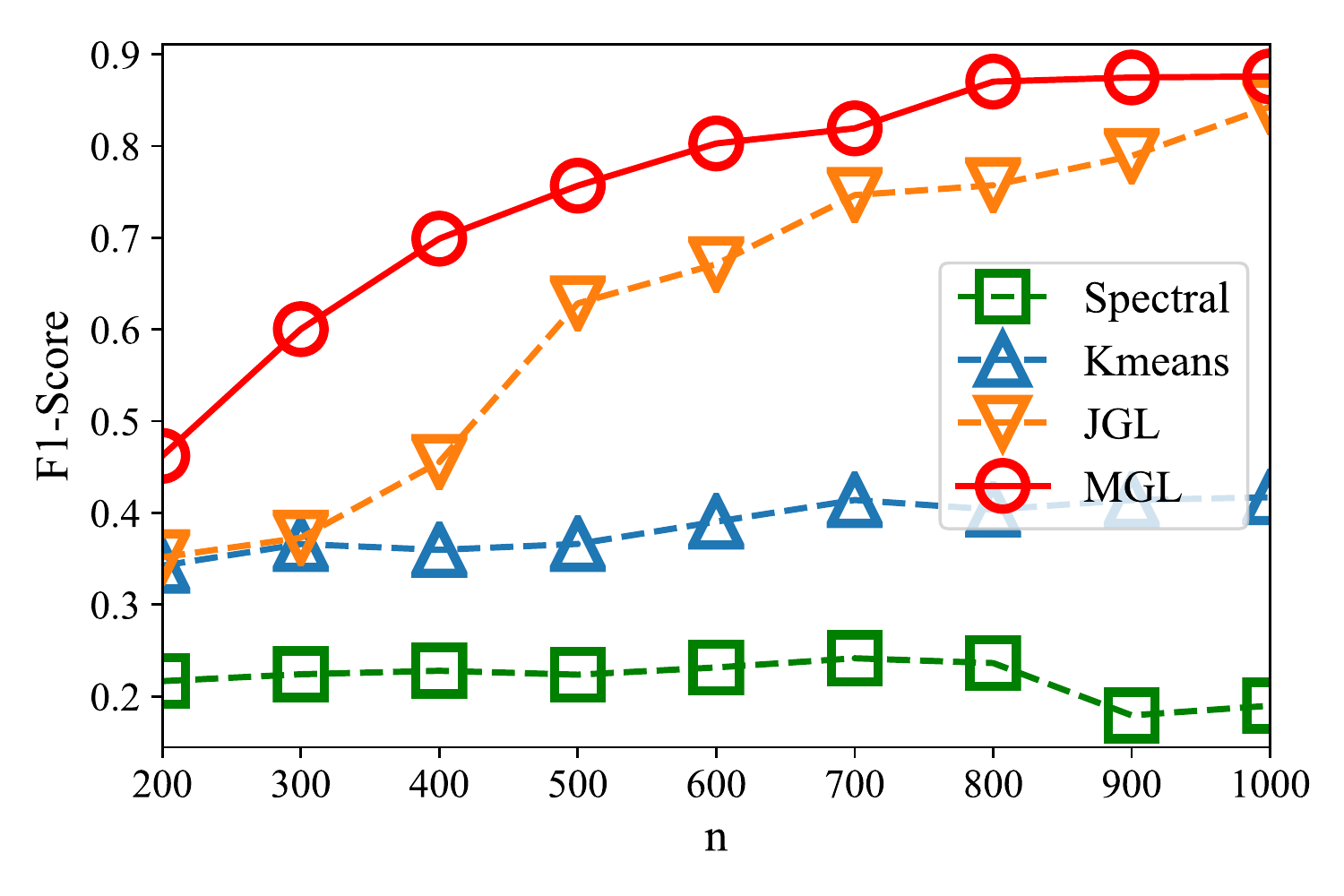}
  \caption{High Dimension - Sample Size (Scenario 3)}
\end{subfigure}
\begin{subfigure}{.45\columnwidth}
\centering
  \includegraphics[width=1\linewidth]{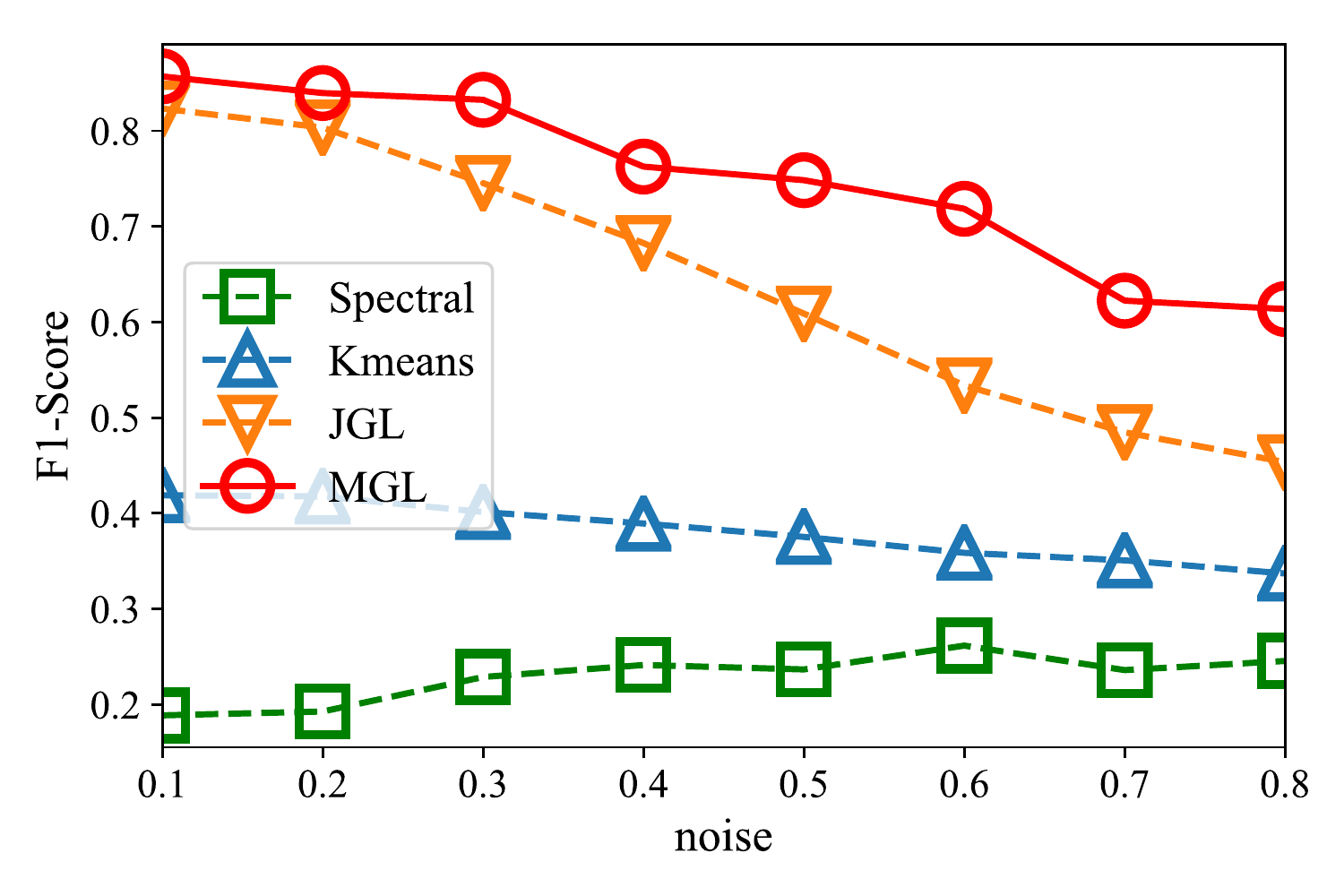}
  \caption{High Dimension - Noise (Scenario 4)}
\end{subfigure}
\caption{Comparison of each model on edge detection. Each figure shows the results of F1-score. The dark blue line indicates GLasso + Spectral; the light blue indicates $k$-means + GLasso; the orange one shows the result of MGL without Mutual Exclusivity Regularization and the green one shows the result of MGL. }
\label{fig:f1-score}
\vspace{5pt}
\end{figure}